\documentclass[10pt,twocolumn,letterpaper]{article}

\usepackage[pagenumbers]{cvpr} 

\usepackage{color, colortbl}
\usepackage{graphicx}
\usepackage{amsmath}
\usepackage{amssymb}
\usepackage{multirow}
\usepackage{booktabs}
\usepackage{threeparttable}
\usepackage{bm}

\definecolor{cvprblue}{rgb}{0.21,0.49,0.74}
\usepackage[pagebackref,breaklinks,colorlinks,allcolors=cvprblue]{hyperref}

\def\confName{CVPR}
\def\confYear{2026}

\title{Parameter-efficient Prompt Tuning and Hierarchical Textual Guidance for Few-shot Whole Slide Image Classification}

\author{Jayanie Bogahawatte, Sachith Seneviratne, Saman Halgamuge \\
AI, Optimization and Pattern Recognition Research Group \\
Dept. of Mechanical Eng., University of Melbourne, Australia\\
{\tt\small jbogahawatte@student.unimelb.edu.au, \{sachith.seneviratne, saman\}@unimelb.edu.au}
}

\begin{document}
\maketitle
\begin{abstract}

Whole Slide Images (WSIs) are giga-pixel in scale and are typically partitioned into small instances in WSI classification pipelines for computational feasibility. However, obtaining extensive instance level annotations is costly, making few-shot weakly supervised WSI classification (FSWC) crucial for learning from limited slide-level labels. Recently, pre-trained vision-language models (VLMs) have been adopted in FSWC, yet they exhibit several limitations. Existing prompt tuning methods in FSWC substantially increase both the number of trainable parameters and inference overhead. Moreover, current methods discard instances with low alignment to text embeddings from VLMs, potentially leading to information loss. To address these challenges, we propose two key contributions. First, we introduce a new parameter efficient prompt tuning method by scaling and shifting features in text encoder, which significantly reduces the computational cost. Second, to leverage not only the pre-trained knowledge of VLMs, but also the inherent hierarchical structure of WSIs, we introduce a WSI representation learning approach with a soft hierarchical textual guidance strategy without utilizing hard instance filtering. Comprehensive evaluations on pathology datasets covering breast, lung, and ovarian cancer types demonstrate consistent improvements up-to 10.9\%, 7.8\%, and 13.8\% respectively, over the state-of-the-art methods in FSWC. Our method reduces the number of trainable parameters by 18.1\% on both breast and lung cancer datasets, and 5.8\% on the ovarian cancer dataset, while also excelling at weakly-supervised tumor localization.

\end{abstract}    
\section{Introduction}
\label{sec:intro}

Computational pathology (CPath) has undergone a paradigm shift with the integration of deep learning methods \cite{mil-survery1, mil-survey2}, enabling automated analysis of Whole Slide Images (WSIs). However, WSIs pose unique challenges compared to natural images due to their gigapixel scale and the confinement of diagnostically relevant features, such as malignant cells to extremely small regions. Therefore, constructing large-scale WSI datasets with precise annotations of those malignant cell regions demands labor-intensive domain expertise \cite{mil-survey3}. To reduce annotation costs, multiple instance learning (MIL) has become a predominant mechanism for WSI classification by training models in a weakly-supervised manner with only slide-level labels \cite{abmil, dsmil, dtfd-mil, trans-mil, annotationeffmil, camil}. MIL-based methods first partition WSIs into smaller patches called instances and treat each slide as a bag of instances without explicitly considering the spatial structure of the WSIs. Then, instance embeddings are obtained using pre-trained encoder models, which are later aggregated into a slide-level or bag-level representation. Despite their success, they require large volumes of slide-level annotated WSIs, which are scarce due to the data sensitivity in medical domain and limited availability for rare diseases \cite{cpath-survey1}. In response to these constraints, few-shot learning frameworks have emerged as a promising direction \cite{fewshot1, top}. In few-shot weakly supervised WSI classification (FSWC), introduced by \cite{top}, only a limited number of slide-level (bag-level) labels per class are required for training.

Recent Vision-Language Models (VLMs) such as CLIP \cite{clip}, ALIGN \cite{align}, and FILIP \cite{filip}, have demonstrated strong zero-shot and few-shot performance on natural imaging tasks, enabled by their transfer learning capabilities \cite{vlm-survey}. This progress has opened promising avenues for leveraging VLMs in CPath, particularly for FSWC. With the availability of CPath-VLMs pretrained on millions of instance level image-text pairs \cite{conch, plip, quilt, benchmarkFM}, transferring rich visual-language representations to downstream tasks with minimal supervision has become feasible. Recent FSWC approaches leverage a prompt tuning strategy known as context optimization (CoOp) \cite{coop}, to adapt VLMs by optimizing a small set of prompt parameters while keeping the pre-trained VLM backbone frozen. To further enhance the textual prompt quality, Large Language Models (LLMs) are employed to generate class-specific descriptions that capture morphological features of different cancer subtypes \cite{top, vila-mil, focus}. However, a core challenge lies in using VLMs pre-trained with instance level image-text pairs, with slide-level representations in the downstream task. Since WSIs are in giga-pixel scale and contain diverse instances, bridging the distribution and semantic gap between localized instance-level features and global slide-level labels is non-trivial. To address this, prior work employs either cross-attention mechanisms \cite{focus, vila-mil} or contrastive learning frameworks \cite{top, fast} together with CoOp-based prompt learning \cite{coop}. While cross-attention offers fine-grained alignment, it introduces a substantial number of trainable parameters, increasing the risk of overfitting, particularly evident in extreme few-shot settings. In addition, CoOp-based contrastive learning methods often struggle to adapt VLMs to slide-level semantics in few-shot regimes and incur additional inference overhead \cite{ssf}.

On the other hand, WSIs possess an inherent hierarchical spatial structure, where fine-grained regions reflect localized cellular interactions and larger regions encode spatial organization of cell clusters and tumor-immune interfaces \cite{hipt}. Since diagnostically relevant regions are spatially confined, aggregating all instance embeddings without considering the spatial structure of WSIs can dilute informative signals \cite{camil}. Moreover, prior methods often discard instances solely based on their alignment with text embeddings obtained from the VLM text encoder to reduce computational cost \cite{focus, mscpt, MAPLE}. In some cases, this results in eliminating over 60\% of the instances in WSIs. However, we suggest that such hard filtering may discard regions with some diagnostically relevant information that may not be represented in the text embeddings. This can be detrimental in few-shot settings as demonstrated through experiments. These limitations can collectively lead to inaccurate tumor localization and thereby compromising the diagnostic reliability. Hence, effective adaptation of VLMs in few-shot settings necessitates a soft textual guidance from the text embeddings from VLM text encoder that integrates the pre-trained knowledge of the VLM together with the intrinsic hierarchical and spatial structure of WSIs during representation learning.

To address the above mentioned limitations, we propose \textbf{H}ierarchical textual guidance based WS\textbf{I} representation learning and \textbf{P}rompt tuning with \textbf{S}caling and \textbf{S}hifting features for FSWC (HIPSS). HIPSS is built upon two key contributions. First, we introduce a new prompt tuning approach, which is inspired by Scaling and Shifting Features (SSF) in pre-trained vision encoders \cite{ssf, fewshotdiscriminative}. We propose to employ a parameter-efficient linear transformation of features in the pre-trained text encoder which significantly reduces the number of trainable parameters as opposed to prior work that utilizes cross attention for feature-fusion \cite{focus, vila-mil, mscpt}. In contrast to CoOp-based methods \cite{coop}, our prompt learning method does not incur additional cost at inference as the linear transformations can be merged into the model using re-parameterization \cite{ssf}. To the best of our knowledge, scaling and shifting features is unexplored in the domain of VLM-based prompt tuning. Second, we propose a new hierarchical textual guidance strategy for WSI representation learning. Each WSI is first partitioned into regions, each of which is further divided into instances. To provide the hierarchical textual guidance, we propose to use attention pooling blocks at each stage that aggregates features together with attention weights refinement using the text embeddings from the VLM text encoder without any hard instance filtering. The main contributions of our work are summarized as follows:
\begin{enumerate}
    \item We introduce a new parameter efficient prompt tuning approach with scaling and shifting features in text encoder.
    \item We introduce a new soft hierarchical textual guidance-based WSI representation learning method.
    \item HIPSS consistently outperforms state-of-the-art (SOTA) methods in comprehensive evaluations across three WSI datasets comprising breast, lung and ovarian cancer types under few-shot settings. 
    \item We demonstrate the quality of the learned WSI representations by achieving strong performance in weakly-supervised tumor localization task.
\end{enumerate}
\section{Related Work}

\begin{figure*}[t]
\centering
\includegraphics[width=0.98\textwidth]{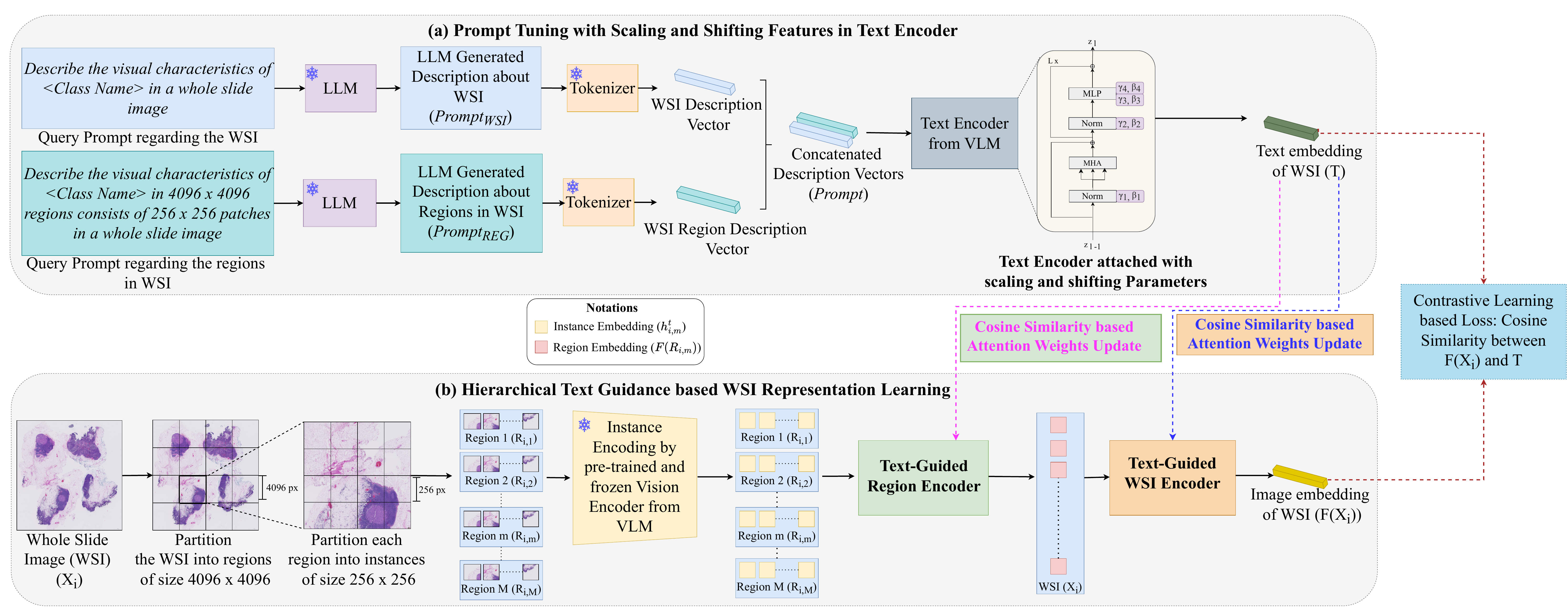}
\caption{Illustration of HIPSS framework. \textbf{(a)}: Two text descriptions are generated considering the entire WSI and the regions in WSI. \textbf{A set of task-specific parameters: \( \gamma\) and \(\beta\) are attached to a set of selected layers of the text encoder backbone, up-to a pre-defined depth}.  \textbf{(b)}: \textbf{WSI representation learning using two attention pooling mechanisms with textual guidance.} Region encoder creates each region embedding considering the instances in the specific region. WSI encoder creates the WSI embedding by aggregating region embeddings. \textbf{Attention weights are refined based on the cosine similarity between instance (or region) embeddings and the text embeddings}. Contrastive learning based loss is calculated between image and text embeddings.}
\label{fig:method.method_main}
\end{figure*}

\subsection{Multiple Instance Learning for WSI Classification}

Multiple Instance Learning (MIL) is a widely adopted weakly supervised approach for WSI classification in CPath, where only slide-level labels are available during training. A key challenge in MIL is to effectively aggregate instance-level features into slide-level representations. Early methods used non-trainable functions (e.g.max/mean pooling), which fail to capture instance importance. Attention-based methods such as ABMIL \cite{abmil} introduced trainable mechanisms to weigh instances by relevance. DTFD-MIL \cite{dtfd-mil} improved localization by forming pseudo-bags and applying attention mechanism. DSMIL \cite{dsmil} incorporated multi-scale features to capture both fine-grained and global patterns. Transformer-based MIL \cite{trans-mil} models leveraged self-attention to model inter-instance dependencies, while self-supervised methods including HIPT \cite{hipt} pre-trains across resolutions but incur high computational costs during pre-training. Spatially-aware MIL methods, including CAMIL \cite{camil}, further enhanced performance via context-driven attention. Despite these advancements, current MIL methods rely heavily on large numbers of labeled WSIs for effective training \cite{top, mi-zero}, which can be a constraint in real-world clinical settings.

\subsection{Few-shot Weakly Supervised WSI Classification (FSWC)}

Few-shot weakly supervised WSI classification presents unique challenges due to the scarcity of slide-level labels and the sparsity of diagnostically relevant regions. Few-shot learning methods in natural image domain often fail under these constraints. TOP \cite{top} introduced a VLM-based framework with instance and bag-level prompt learning. However, effective dataset-specific instance prompt design demands substantial domain expertise. PEMP \cite{pemp} improves alignment via task-specific visual exemplars, though it assumes availability of auxiliary samples, limiting scalability. ViLa-MIL \cite{vila-mil}, MSCPT \cite{mscpt}, and MAPLE \cite{MAPLE} leverage multi-scale vision-language fusion using cross-attention and graph neural networks. Nevertheless, these methods depend on paired WSIs across magnifications, which can be impractical in many settings. FOCUS \cite{focus} introduces adaptive visual token compression to reduce computational overhead; however, this strategy may inadvertently discard subtle yet clinically significant regions. FAST \cite{fast} combines slide and instance-level supervision for enhanced localization.  While this hybrid annotation setting improves discriminative localization and classification, it assumes access to expert-annotated instances, which may constrain scalability in low-resource or rapidly evolving diagnostic contexts.

\noindent\textbf{VLM-based Prompt Learning in FSWC:} Current methods in this space rely on VLMs such as CLIP \cite{clip} and CONCH \cite{conch}, using a prompt learning strategy known as CoOp \cite{coop} to align visual and textual modalities. CoOp introduces learnable context vectors into the prompt embeddings while keeping the image and text encoders frozen, enabling efficient adaptation to new tasks with minimal supervision. Unlike the natural image domain, CPath involves a fundamental distribution shift between WSI instances and slide-level representations. VLMs such as CONCH \cite{conch} are typically pre-trained on WSI instances, which limits their ability to capture global context necessary for WSI classification in few-shot settings. Consequently, existing prompt learning approaches such as CoOp may struggle to adapt these VLMs for slide-level classification or require computational costly fusing techniques including cross-attention \cite{focus, vila-mil}, resulting in suboptimal performance in 1 or 2-shot settings.

\section{Method}
This section outlines the proposed HIPSS method as illustrated in Fig. \ref{fig:method.method_main}. HIPSS consists of a scaling and shifting feature-based prompt tuning approach and a soft hierarchical textual guidance based WSI representation learning strategy.

\subsection{Problem Definition}

Let \( X = {\{X_1, X_2, ..., X_N\}}\) be a dataset of \(N\) WSIs, where \(X_i\) denotes a WSI which is associated with a slide-level label \(Y_i\). The objective is to minimize the error between predicted slide-level label and ground truth label, under weak supervision without any instance-level labels. In FSWC setting, there is only a limited number of labeled WSIs available during training. We adopt a \(k\)-shot formulation, in which only \(k\) number of labeled WSIs per class are available for training.

\subsection{Pathology Query Prompt Formulation}
\label{subsec: promptFormulation}

CPath-VLMs are typically pre-trained on WSI instance level image-text pairs \cite{conch, plip, quilt}. To leverage these VLMs for FSWC, it is essential to formulate the textual prompts that effectively represent the visual characteristics of WSIs. Therefore, we begin by formulating a structured query to use with a LLM, to generate a detailed textual description of the WSI. Following the hierarchical structure of the WSIs, we design two queries to obtain two prompts containing visual characteristics of the WSIs at region level (\(Prompt_{REG}\)) and at slide level (\(Prompt_{WSI}\)). This formulation incorporates both fine-grained region or instance-level attributes and slide-level context. The LLM-generated descriptions are then passed through the tokenizer in the VLM to obtain the two token sequences corresponding to each prompt. These two token sequences are concatenated to use as the input to the text encoder in VLM, which is denoted as \(Prompt\) in Fig. \ref{fig:method.method_main} (a).

\subsection{Prompt Tuning with Scaling and Shifting Features in Text Encoder}
\label{subsec: prompt}

Since the VLMs are pre-trained on WSI instance level image-text pairs, the text encoder often require finetuning to better handle target WSI downstream tasks. To adapt the pre-trained text features for WSI classification task, we introduce scaling and shifting features (SSF) in text encoder as the prompt tuning mechanism. Specifically, for each selected layer in the text encoder, a pair of task-specific learnable scale and shift parameters, \(\gamma \in \mathbb{R}^{D_t}\) and \(\beta \in \mathbb{R}^{D_t}\) are attached. Features of each layer, \(x \in \mathbb{R}^{D_t}\) are transformed into \(y \in \mathbb{R}^{D_t}\) using \(\gamma\) and \(\beta\), as defined by Eq. \ref{eq:method.ssf}, where \(\cdot\) is the dot product. Here, \(D_t\) is the dimension of the text embeddings in VLM. During training, the pre-trained weights of the layers in text encoder are kept frozen and only the  \(\gamma\) and \(\beta\) parameters are updated. This approach is motivated by alignment of feature distributions, where only two parameters per layer are learned to modulate pre-trained features to the target distribution.
\begin{equation}
\label{eq:method.ssf}
y = \gamma \cdot x + \beta
\end{equation}
\noindent\textbf{Tuning depth of text encoder using SSF:} To control the degree of adaptation, we selectively insert the SSF parameters only to a subset of blocks in the text encoder by varying the depth or number of trainable blocks, \(d_s\). The list of blocks which we apply SSF is defined as $[(L - d_s + 1), ..., L]$, where \(L\) is the total number of blocks in the text encoder.

\noindent We tune different types of layers in each block in the text encoder using SSF, including layer normalization and multi layer perceptron as depicted in Fig. \ref{fig:method.method_main} (a). While the types of layers that we attach the SSF parameters can be customized, our implementation adopts SSF tuning across these layers to maximize adaptability while reducing computational cost. \(Prompt\) is passed through the text encoder with SSF tuning to generate the text embeddings \(T\).

\subsection{Hierarchical Textual Guidance based WSI Representation Learning}

In hierarchical textual guidance based WSI representation learning, region-level embeddings are computed using a region encoder, while the WSI encoder aggregates region embeddings to generate the final slide-level representation. Attention weights at each stage are refined using the text embeddings obtained from VLM text encoder to provide the soft textual guidance in a hierarchical manner. 

\noindent\textbf{Region Encoder:} As depicted in Fig. \ref{fig:method.method_main} (b), we first partition each WSI, \(X_i\) into regions of size \(4096 \times 4096\). Each WSI is represented as, \( X_i = {\{R_{i,1}, R_{i,2}, ..., R_{i,M}\} }\), where \(M\) denotes the number of regions in slide \(X_i\), which varies across slides. Secondly, each region \(R_{i,m}\) is further partitioned into instances of size \(256 \times 256\). Each region can be denoted as \( R_{i,m} = {\{P_{i,m}^1, P_{i,m}^2, ..., P_{i,m}^J\} }\), where \(J\) is the number of instances in \(R_{i,m}\). To extract the feature representations of the instances, we use the pre-trained image encoder of the VLM while keeping its weights frozen during training. Let \(h_{i,m}^j \in \mathbb{R}^{D_v}\) be the extracted feature representation of an instance \(P_{i,m}^j\). \(D_v\) is the dimension of the feature vector, which is same as the feature dimension of the VLM image encoder. Feature representation \(F(R_{i,m})\) of a region \(R_{i,m}\) is computed using an attention pooling block denoted by Eq. \ref{eq:method.region_repre}, only considering the instances in the \(R_{i,m}\) region:
\begin{equation}
\label{eq:method.region_repre}
F(R_{i,m}) = \sum\limits_{j=1}^J a_{i,m}^j h_{i,m}^j \in \mathbb{R}^{D_v},
\end{equation}
where \(a_{i,m}^j\) is the learnable attention weight for \(h_{i,m}^j\), which is determined by Eq. \ref{eq:method.attention_weight_region}, following \cite{abmil} and modified using the proposed text-guided attention refinement score denoted by \(s_{i,m}^j\):
\begin{equation}
\hspace{-1.0em}
\label{eq:method.attention_weight_region}
a_{i,m}^j = \frac{\exp \{w_r^\top (\tanh(V_1{h_{i,m}^j}^\top) \odot \sigma(V_2{h_{i,m}^j}^\top)) +  s_{i,m}^j \}} {\sum\limits_{k=1}^T \exp \{w_r^\top (\tanh(V_1{h_{i,m}^k}^\top) \odot \sigma(V_2{h_{i,m}^k}^\top)) + s_{i,m}^k\}},
\end{equation}
Here, \(w_r\), \(V_1\), and \(V_2\) are learnable parameters, \(\odot\) is the element-wise multiplication, and \(\sigma\) denotes the sigmoid function. \(s_{i,m}^j\) is the score which is computed to refine the attention weights using the text embeddings obtained from the VLM text encoder, using Eq. \ref{eq:method.score_patch}:
\begin{equation}
\hspace{-1.5em}
\label{eq:method.score_patch}
s_{i,m}^j = \begin{cases}
        \lambda * cos(h_{i,m}^j, T), & \text{if } cos(h_{i,m}^j, T) >\alpha \\
        cos(h_{i,m}^j, T), & \text{if } 0 <cos(h_{i,m}^j, T) <\alpha \\
        0, & \text{otherwise}
    \end{cases}
\end{equation}
where \(cos(.,.)\) is the cosine similarity between feature representation of each instance in a specific region and the text embedding (\(T\)) obtained from VLM text encoder (Refer Sec. \ref{subsec: promptFormulation} and \ref{subsec: prompt}). For binary classification tasks, \(T\) corresponds to the tumor class description embedding and for multi-cancer subtype classification, \(T\) is computed as the mean embedding of all subtype descriptions, providing a unified semantic representation of cancer phenotypes. $\lambda$ and $\alpha$ are the empirically determined refinement factor and threshold values respectively. We formulate soft textual guidance with three alignment regimes based on cosine similarity. Negative similarities are suppressed as semantically inconsistent; low positive similarities are linearly weighted to retain uncertain but informative instances; and similarities above a threshold are amplified to emphasize high-confidence semantic matches. This enables a reliability-aware gating and weighting mechanism without hard instance filtering.

\noindent\textbf{WSI Encoder:} To compute the feature representation of the WSI \(F(X_i)\), we use a different attention-pooling block as formulated by Eq. \ref{eq:method.wsi_repre} using the region embeddings computed with Eq. \ref{eq:method.region_repre}:
\begin{equation}
\label{eq:method.wsi_repre}
F(X_i) = \sum\limits_{m=1}^M a_{i,m} (F(R_{i,m})) \in \mathbb{R}^{D_v},
\end{equation}
where \(a_{i,m}\) is the learnable attention weight for \(F(R_{i,m})\), which is determined by Eq. \ref{eq:method.attention_weight_wsi}. \(F(R_{i,m})\) is denoted by \(r_{i,m}\) in Eq. \ref{eq:method.attention_weight_wsi} for simplicity.
\begin{equation}
\label{eq:method.attention_weight_wsi}
a_{i,m} = \frac{\exp \{w^\top (\tanh(U_1{r_{i,m}}^\top) \odot \sigma(U_2{r_{i,m}}^\top)) + s_{i,m}\}} {\sum\limits_{k=1}^M \exp \{w^\top (\tanh(U_1{r_{i,k}}^\top)) \odot \sigma(U_2{r_{i,k}}^\top)) + s_{i,k}\}},
\end{equation}
In Eq. \ref{eq:method.attention_weight_wsi}, \(w\), \(U_1\), and \(U_2\) are learnable parameters, \(\odot\) is the element-wise multiplication, \(\sigma\) is the sigmoid function and \(s_{i,m}\) is the score computed similar to Eq. \ref{eq:method.score_patch} for each region. Finally, the feature representation of the WSI, \(F(X_i)\) is later used in the classification.

\subsection{Contrastive Learning based Training Strategy}

With the utilization of the VLM image encoder and hierarchical textual guidance based WSI representation learning module, the visual embedding of each WSI is generated as \(F(X_i)\). The text descriptions are passed through the text encoder with SSF tuning to generate the textual embeddings \(T\). These embeddings are jointly optimized using a contrastive loss that brings paired image-text representations closer in the shared embedding space and repels non-matching pairs \cite{top}. The predicted probability for class \(c\) is defined by Eq. \ref{eq:method.contrastive}:
\begin{equation}
\label{eq:method.contrastive}
P(Y_i=c| X_i) = \frac{\exp (cos (F(X_i), T_c) /  \tau) } {\sum\limits_{j=1}^C \exp (cos (F(X_i), T_j)) /  \tau )},
\end{equation}
where \(cos(.,.)\) is the cosine similarity between embeddings and each embedding is normalized, \(C\) is the number of classes, and \(\tau\) is the temperature coefficient. We utilize the cross entropy loss between the predicted probability, \(P(X_i)\) and ground truth slide-level label, \(Y_i\) for the end-to-end training of HIPSS pipeline.

\section{Experiments and Results}

The proposed method is evaluated on FSWC tasks using three distinct datasets, each representing a different cancer type. We conducted experiments under varying few-shot settings, utilizing 1, 2, 4, 8, and 16 labeled WSIs per class for training, referred to as 1-shot, 2-shot, 4-shot, 8-shot, and 16-shot settings respectively.

\begin{table*}[t]
\centering
\fontsize{9pt}{9pt}\selectfont
\caption{Few Shot Classification results for Camelyon16, TCGA-Lung, and UBC-OCEAN Datasets. Mean AUC values averaged over 3 different folds, each evaluated using 20 random seeds are reported (Refer to Section \ref{sec:experimental_details} for further details). Performance improvement compared to the second best method is reported within brackets with (+). Bold: best results, Underline: second best results.}
\centering
\begin{tabular}{c|lccccc}
\toprule
{Dataset} & {Method}   & {16-shot setting} & {8-shot setting} & {4-shot setting} & {2-shot setting} & {1-shot setting}\\
\midrule
\multirow{11}{*}{\rotatebox{90}{Camelyon16}}
	 & Max Pooling - CLIP	 & 	$0.5947_{\pm 0.0135}$	 & 	$0.5814_{\pm 0.0112}$	 & 	$0.5417_{\pm 0.0126}$ & 	$0.5401_{\pm 0.0197}$	 & 	$0.5268_{\pm 0.0154}$ \\
     & Mean Pooling - CLIP	 & 	$0.5816_{\pm 0.0127}$	 & 	$0.5627_{\pm 0.0098}$	 & 	$0.4930_{\pm 0.0122}$ & 	$0.3235_{\pm 0.0178}$	 & 	$0.3357_{\pm 0.0173}$ \\
	 & Max Pooling	& 	$0.6709_{\pm 0.0182}$	 & 	$0.6725_{\pm 0.0095}$	 & 	$0.6263_{\pm 0.0128}$ & 	$0.5684_{\pm 0.0406}$	 & 	$0.5732_{\pm 0.0191}$ \\
	 & Mean Pooling	 & 	$0.6843_{\pm 0.0193}$	 & 	$0.6330_{\pm 0.0449}$	 & 	$0.5959_{\pm 0.0141}$ & 	$0.5637_{\pm 0.0318}$	 & 	$0.5716_{\pm 0.0259}$ \\
	 & ABMIL (ICML' 18) 	 & 	$0.7820_{\pm 0.0455}$	 & 	$0.7195_{\pm 0.0314}$	 & 	$0.6086_{\pm 0.0587}$ & 	$0.5969_{\pm 0.0431}$	 & 	$0.5706_{\pm 0.0426}$ \\
     & ViLa-MIL (CVPR' 24)  & 	$0.7997_{\pm 0.0383}$	 & 	$0.7599_{\pm 0.0474}$	 & 	$0.5617_{\pm 0.0543}$ & 	$0.4601_{\pm 0.0365}$	 & 	$0.4565_{\pm 0.0280}$ \\
	 & TOP (NeurIPS' 23)  & $0.8345_{\pm 0.0167}$	 & 	$0.7301_{\pm 0.0133}$	 & 	$0.7282_{\pm 0.0132}$ & 	$0.7054_{\pm 0.0179}$	 & 	$0.6852_{\pm 0.0199}$ \\
    & FAST (NeurIPS' 24)  & 	$0.8197_{\pm 0.0474}$	 & 	$0.7742_{\pm 0.0249}$	 & 	\underline{$0.7359_{\pm 0.0853}$} & 	\underline{$0.7595_{\pm 0.0391}$}	 & 	\underline{$0.6933_{\pm 0.0846}$} \\
     & FOCUS (CVPR' 25) 	 & \underline{$0.8402_{\pm 0.0412}$}	 & 	\underline{$0.7974_{\pm 0.0499}$}	 & 	$0.6366_{\pm 0.0555}$ & 	$0.5682_{\pm 0.0117}$	 & 	$0.5265_{\pm 0.0224}$ \\
     \cmidrule{2-7}
 & \cellcolor{gray!20!white} HIPSS (Ours) & 	\cellcolor{gray!20!white} $\bm{0.9105_{\pm 0.0104}}$	&  \cellcolor{gray!20!white} $\bm{0.8586_{\pm 0.0067}}$ &	\cellcolor{gray!20!white} $\bm{0.8159_{\pm 0.0165}}$  &		\cellcolor{gray!20!white} $\bm{0.8066_{\pm 0.0069}}$  &	\cellcolor{gray!20!white} $\bm{0.7429_{\pm 0.0087}}$ \\
  &  & 	\textcolor{teal}{(+8.4)}	&  \textcolor{teal}{(+7.7)} &	\textcolor{teal}{(+10.9)}  &		\textcolor{teal}{(+6.2)}  &	\textcolor{teal}{(+7.2)} \\
\midrule
\multirow{10}{*}{\rotatebox{90}{TCGA-Lung}}
	 & Max Pooling - CLIP	 & 	$0.6227_{\pm 0.0092}$	 & 	$0.5547_{\pm 0.0137}$	 & 	$0.5155_{\pm 0.0128}$ & 	$0.4985_{\pm 0.0103}$	 & 	$0.4876_{\pm 0.0087}$ \\
     & Mean Pooling - CLIP	 & 	$0.6022_{\pm 0.0083}$	 & 	$0.5418_{\pm 0.0109}$	 & 	$0.4934_{\pm 0.0167}$ & 	$0.4908_{\pm 0.0078}$	 & 	$0.4876_{\pm 0.0099}$ \\
	 & Max Pooling	& 		$0.7781_{\pm 0.0067}$	 & 	$0.7009_{\pm 0.0133}$	 & 	$0.6129_{\pm 0.0264}$ & 	$0.6301_{\pm 0.0156}$	 & 	$0.6007_{\pm 0.0141}$ \\
	&  Mean Pooling	 & 	$0.7618_{\pm 0.0123}$	 & 	$0.7409_{\pm 0.0099}$	 & 	$0.6357_{\pm 0.0141}$ & 	$0.6341_{\pm 0.0164}$	 & 	$0.5889_{\pm 0.0194}$ \\
	 & ABMIL (ICML' 18)	 & 	$0.8074_{\pm 0.0166}$	 & 	$0.7709_{\pm 0.0118}$	 & 	$0.6832_{\pm 0.0281}$ & 	$0.6541_{\pm 0.0166}$	 & 	$0.5965_{\pm 0.0297}$ \\
     & ViLa-MIL (CVPR' 24) & 	$0.8787_{\pm 0.0176}$	 & 	$0.8012_{\pm 0.0191}$	 & 	$0.6942_{\pm 0.0248}$ & 	$0.6385_{\pm 0.0472}$	 & 	$0.5761_{\pm 0.0345}$ \\
	 & TOP (NeurIPS' 23)  & $0.8256_{\pm 0.0189}$	 & 	$0.8161_{\pm 0.0181}$	 & 	\underline{$0.7689_{\pm 0.0132}$} & 	\underline{$0.7341_{\pm 0.0178}$}	 & 	\underline{$0.7131_{\pm 0.0162}$} \\
     & FOCUS (CVPR' 25)	 & 	$0.8717_{\pm 0.0237}$	 & 	$0.7688_{\pm 0.0381}$	 & 	$0.7111_{\pm 0.0385}$ & 	$0.6515_{\pm 0.0681}$	 & 	$0.5665_{\pm 0.0455}$ \\
      & MAPLE (NeurIPS' 25)	 & 	\underline{$0.9044_{\pm 0.0128}$}	 & 	\underline{$0.8579_{\pm 0.0221}$}	 & 	$0.7558_{\pm 0.0161}$ & 	$0.7213_{\pm 0.0284}$	 & 	$0.6995_{\pm 0.0216}$ \\
\cmidrule{2-7}
 & \cellcolor{gray!20!white} HIPSS (Ours) & 	\cellcolor{gray!20!white} $\bm{0.9097_{\pm 0.0155}}$	&  \cellcolor{gray!20!white} $\bm{0.8745_{\pm 0.0312}}$ &	\cellcolor{gray!20!white} $\bm{0.8285_{\pm 0.0205}}$  &		\cellcolor{gray!20!white} $\bm{0.7616_{\pm 0.0181}}$  &	\cellcolor{gray!20!white} $\bm{0.7217_{\pm 0.0141}}$ \\
   &  & 	\textcolor{teal}{(+0.6)}	&  \textcolor{teal}{(+1.9)} &	\textcolor{teal}{(+7.8)}  &		\textcolor{teal}{(+3.7)}  &	\textcolor{teal}{(+1.2)} \\
\midrule
\multirow{10}{*}{\rotatebox{90}{UBC-OCEAN}}

 	 & Max Pooling - CLIP	 & 	$0.8012_{\pm 0.0082}$	 & 	$0.8127_{\pm 0.0091}$	 & 	$0.7056_{\pm 0.0069}$ & 	$0.5123_{\pm 0.0071}$	 & 	$0.4926_{\pm 0.0095}$ \\
     &  Mean Pooling - CLIP	 & 	$0.8001_{\pm 0.0061}$	 & 	$0.8068_{\pm 0.0077}$	 & 	$0.6997_{\pm 0.0051}$ & 	$0.5154_{\pm 0.0062}$	 & 	$0.4818_{\pm 0.0083}$ \\
	 & Max Pooling	& 	$0.8496_{\pm 0.0087}$	 & 	$0.8373_{\pm 0.0060}$	 & 	$0.7922_{\pm 0.0087}$ & 	$0.5574_{\pm 0.0237}$	 & 	$0.5139_{\pm 0.0204}$ \\
	 & Mean Pooling	 &  $0.8456_{\pm 0.0064}$	 & 	$0.8394_{\pm 0.0056}$	 & 	$0.7910_{\pm 0.0081}$ & 	$0.6609_{\pm 0.0214}$	 & 	$0.5990_{\pm 0.0192}$ \\
	 & ABMIL (ICML' 18)	 & 	$0.8432_{\pm 0.0132}$	 & 	$0.8425_{\pm 0.0068}$	 & 	$0.7857_{\pm 0.0107}$ & 	$0.7037_{\pm 0.0143}$	 & 	$0.6244_{\pm 0.0237}$ \\
     & ViLa-MIL (CVPR' 24) & 	$0.9199_{\pm 0.0047}$	 & 	$0.9051_{\pm 0.0086}$	 & 	$0.8489_{\pm 0.0049}$ & 	$0.7753_{\pm 0.0150}$	 & 	$0.7212_{\pm 0.0141}$ \\
	 & TOP (NeurIPS' 23)	 &  $0.8003_{\pm 0.0022}$	 & 	$0.7560_{\pm 0.0048}$	 & 	$0.7295_{\pm 0.0274}$ & 	$0.6272_{\pm 0.0014}$	 & 	$0.5449_{\pm 0.0291}$ \\
     & FOCUS (CVPR' 25)	 & 	\underline{$0.9433_{\pm 0.0034}$}	 & 	\underline{$0.9394_{\pm 0.0051}$}	 & 	\underline{$0.9144_{\pm 0.0065}$} & 	\underline{$0.8487_{\pm 0.0047}$}	 & 	\underline{$0.7508_{\pm 0.0091}$} \\
\cmidrule{2-7}
 & \cellcolor{gray!20!white} HIPSS (Ours) & 	\cellcolor{gray!20!white} $\bm{0.9540_{\pm 0.0138}}$	&  \cellcolor{gray!20!white} $\bm{0.9516_{\pm 0.0121}}$ &	\cellcolor{gray!20!white} $\bm{0.9305_{\pm 0.0131}}$  &		\cellcolor{gray!20!white} $\bm{0.8797_{\pm 0.0152}}$  &	\cellcolor{gray!20!white} $\bm{0.8542_{\pm 0.0146}}$ \\
 &  & 	\textcolor{teal}{(+1.1)}	&  \textcolor{teal}{(+1.3)} &	\textcolor{teal}{(+1.8)}  &		\textcolor{teal}{(+3.7)}  &	\textcolor{teal}{(+13.8)} \\
\bottomrule
\end{tabular}
\fontsize{9pt}{9pt}\selectfont
\label{tab:exp.3datasets}
\end{table*}

\begin{table}[t]
\caption{Ablation Studies on 3 datasets. Mean AUC values are reported. Components are denoted as H: Hierarchical WSI representation learning, T: Text-guided attention refinement strategy, and S: SSF-based prompt tuning in text encoder. Full results are provided in the Appendix.}
\centering
\resizebox{0.95\columnwidth}{!}{
\centering
\begin{tabular}{ccccccc}
\toprule[1.0pt]
Dataset & H & T & S & 16-shot & 4-shot& 1-shot  \\
\midrule[1.0pt]
\multirow{6}{*}{\rotatebox{90}{Camelyon16}}
& $\times$ & $\times$ & $\times$ & $0.7820_{\pm 0.0455}$	  & 	$0.6086_{\pm 0.0587}$ 	 & 	$0.5706_{\pm 0.0426}$ \\
& $\checkmark$ & $\times$ & $\times$ & $0.8666_{\pm 0.0085}$		 & 	$0.6903_{\pm 0.0170}$ 	 & 	$0.6433_{\pm 0.0169}$ \\
& $\checkmark$ & $\checkmark$ & $\times$ & $0.8724_{\pm 0.0086}$	 	 & 	$0.7552_{\pm 0.0173}$ 	 & 	$0.6512_{\pm 0.0089}$ \\
& $\times$ & $\times$ & $\checkmark$ & $0.8853_{\pm 0.0267}$	 	 & 	$0.7798_{\pm 0.0033}$ 	 & 	$0.7071_{\pm 0.0070}$ \\
& $\checkmark$ & $\times$ & $\checkmark$ & $0.9006_{\pm 0.0116}$	 	 & 	$0.8034_{\pm 0.0139}$ 	 & 	$0.7217_{\pm 0.0081}$ \\
& \cellcolor{gray!20!white} $\checkmark$ & \cellcolor{gray!20!white} $\checkmark$ & \cellcolor{gray!20!white} $\checkmark$ & \cellcolor{gray!20!white} $\bm{0.9105_{\pm 0.0104}}$ &	\cellcolor{gray!20!white} $\bm{0.8159_{\pm 0.0165}}$  &	\cellcolor{gray!20!white} $\bm{0.7429_{\pm 0.0087}}$ \\

\midrule[1.0pt]
\multirow{6}{*}{\rotatebox{90}{TCGA-Lung}}
& $\times$ & $\times$ & $\times$ & $0.8074_{\pm0.0166}$	 	 & 	$0.6832_{\pm 0.0281}$	 & 	$0.5965_{\pm 0.0297}$ \\
& $\checkmark$ & $\times$ & $\times$ & $0.8487_{\pm 0.0305}$		 & 	$0.7420_{\pm 0.0148}$  & 	$0.6202_{\pm 0.0197}$ \\
& $\checkmark$ & $\checkmark$ & $\times$ & $0.8661_{\pm 0.0154}$	 	 & 	$0.7451_{\pm 0.0093}$	 & 	$0.6311_{\pm 0.0123}$ \\
& $\times$ & $\times$ & $\checkmark$ & $0.8781_{\pm 0.0469}$	 	 & 	$0.7757_{\pm 0.0098}$ 	 & 	$0.6801_{\pm 0.0197}$ \\
& $\checkmark$ & $\times$ & $\checkmark$ & $0.9072_{\pm 0.0051}$	 	 & 	$0.7842_{\pm 0.0029}$ 	 & 	$0.7203_{\pm 0.0034}$ \\
& \cellcolor{gray!20!white} $\checkmark$ & \cellcolor{gray!20!white} $\checkmark$ & \cellcolor{gray!20!white} $\checkmark$ & \cellcolor{gray!20!white} $\bm{0.9097_{\pm 0.0155}}$ &	\cellcolor{gray!20!white} $\bm{0.8285_{\pm 0.0205}}$   &	\cellcolor{gray!20!white} $\bm{0.7217_{\pm 0.0141}}$ \\

\midrule[1.0pt]
\multirow{6}{*}{\rotatebox{90}{UBC-OCEAN}}
& $\times$ & $\times$ & $\times$ & $0.8432_{\pm 0.0132}$	 & 	$0.7857_{\pm 0.0107}$ 	 & 	$0.6244_{\pm 0.0237}$ \\
& $\checkmark$ & $\times$ & $\times$ & $0.9165_{\pm 0.0064}$	 & 	$0.8095_{\pm 0.0068}$ 	 & 	$0.6885_{\pm 0.0089}$ \\
& $\checkmark$ & $\checkmark$ & $\times$ & $0.9347_{\pm 0.0074}$		 & 	$0.8233_{\pm 0.0101}$ 	 & 	$0.6983_{\pm 0.0132}$ \\
& $\times$ & $\times$ & $\checkmark$ & $0.9467_{\pm 0.0036}$		 & 	$0.8330_{\pm 0.0032}$ 	 & 	$0.8278_{\pm 0.0054}$ \\
& $\checkmark$ & $\times$ & $\checkmark$ & $0.9517_{\pm 0.0023}$	 	 & 	$0.9320_{\pm 0.0056}$ 	 & 	$0.8498_{\pm 0.0019}$ \\
& \cellcolor{gray!20!white} $\checkmark$ & \cellcolor{gray!20!white} $\checkmark$ & \cellcolor{gray!20!white} $\checkmark$ & \cellcolor{gray!20!white} $\bm{0.9540_{\pm 0.0138}}$ & \cellcolor{gray!20!white}	$\bm{0.9305_{\pm 0.0131}}$  &	\cellcolor{gray!20!white} $\bm{0.8542_{\pm 0.0146}}$ \\

\bottomrule[1.0pt]
\end{tabular}
}
\label{tab:exp.ablations}
\end{table}

\begin{table}[t]
\caption{Performance comparison with LoRA-based prompt tuning on C16 dataset. Mean AUC values are reported. Number of trainable parameters of LoRA-based prompt tuning is 0.7053 M.}
\centering
\resizebox{0.95\columnwidth}{!}{
\centering
\begin{tabular}{cccccc}
\toprule[1.0pt]
Method & 16-shot & 4-shot & 1-shot  \\
\midrule[1.0pt]
LoRA-based prompt tuning & $0.8513_{\pm 0.0162}$		 & 	$0.7668_{\pm 0.0258}$ & 	$0.5667_{\pm 0.0289}$ \\
\cellcolor{gray!20!white} HIPSS (Ours) & 	\cellcolor{gray!20!white} $\bm{0.9105_{\pm 0.0104}}$	&	\cellcolor{gray!20!white} $\bm{0.8159_{\pm 0.0165}}$  &	\cellcolor{gray!20!white} $\bm{0.7429_{\pm 0.0087}}$ \\
\bottomrule[1.0pt]
\end{tabular}
}
\label{tab:exp.lora}
\end{table}

\subsection{Experimental Details}
\label{sec:experimental_details}

\textbf{Datasets:} We perform comprehensive experiments across three publicly available CPath datasets: Camelyon16 \cite{camelyon16}, TCGA-Lung\footnote{\url{https://portal.gdc.cancer.gov}}, and UBC-OCEAN \cite{ubc}, encompassing breast, lung and ovarian cancer types respectively. Camelyon16 (C16) dataset comprises normal slides and slides with breast cancer metastases. TCGA-Lung dataset includes two lung cancer subtypes and UBC-OCEAN is a multi-class dataset, which covers five ovarian cancer subtypes. The WSIs in Camelyon16 and TCGA-Lung datasets were acquired at a magnification level of \(\times40\), while those in UBC-OCEAN dataset were obtained at \(\times20\).

\noindent\textbf{Evaluation Metrics:} Following prior work \cite{top}, we use Area Under the Curve (AUC) as the evaluation metric. To ensure robustness and mitigate the effects of randomness inherent in few-shot settings, we report the mean AUC scores averaged over 3 different folds, each evaluated using 20 random seeds. In accordance with standard evaluation protocols, training and validation folds are constructed for each dataset, while the test sets remain fixed and independent throughout all experiments for each dataset.

\noindent\textbf{Implementation Details:} In accordance with the pre-processing pipeline outlined in CLAM \cite{clam}, we first removed the background areas from the WSI. As the next step, each WSI was segmented into non-overlapping regions of size \(4096 \times 4096\), which were later partitioned into \(256 \times 256\) instances. Since the spatial extent of tumor regions is unknown during training, the appropriate granularity should be empirically determined. We observed that using excessively large regions risks diluting discriminative features with irrelevant context, whereas overly small regions may possess limited receptive fields by the drop in performances with both. Pathology related descriptions were generated using ChatGPT-4o, based on a set of queries for each class. We use the same set of generated textual descriptions across all experiments to ensure consistency. We employed the CONCH model \cite{conch, benchmarkFM} as the VLM. We randomly initialized scale and shift parameters with a mean value of one and mean value of zero respectively. The number of layers that we tune using SSF in the text encoder was determined empirically. 2 layers were used for C16 and TCGA-Lung datasets, while 8 layers were found optimal for UBC-OCEAN dataset for SSF tuning. Attention weights refinement factor was 10 and threshold value was 0.2 through all experiments. Section \ref{sec:ablation_and_analysis} provides more discussion on these choices. We utilized Adam optimizer \cite{adam} with learning rates ranging from 0.01 to 0.0001. We trained our model for 100 epochs with early stopping based on validation performance. All experiments were conducted on a single 80GB Nvidia A100 GPU.

\noindent\textbf{Baselines:} We compare our proposed method with SOTA methods in conventional WSI classification and few-shot weakly supervised WSI classification. Baseline methods encompass a wide range of models including Max/Mean pooling, ABMIL \cite{abmil}, ViLa-MIL \cite{vila-mil}, TOP \cite{top}, FOCUS \cite{focus}, FAST \cite{fast}, and MAPLE \cite{MAPLE}. To ensure fair comparison across all methods, we employed CONCH model as the VLM backbone for all the baseline methods. For the FAST method, we report the results directly from \cite{fast}, as this approach requires instance-level annotations for WSIs during training, which are not publicly available. Consequently, reproducing the method under our experimental setup was not feasible for TCGA-Lung and UBC-OCEAN datasets. For the MAPLE method, we report the results only for TCGA-Lung dataset with the text prompts and settings provided in \cite{MAPLE}. Additionally, we incorporated CLIP-based models \cite{clip} with Max/Mean pooling as baseline methods.

\subsection{Comparison Results}

\textbf{C16 (2 classes):} Table \ref{tab:exp.3datasets} shows that our method consistently outperforms existing SOTA methods across all few-shot settings on C16 dataset, with gains of 7.2\% (1-shot) and 6.2\% (2-shot). HIPSS achieves improvements of 10.9\%, 7.7\%, and 8.4\% in the 4, 8, and 16 shot settings, respectively. 

\noindent\textbf{TCGA-Lung (2 classes):} As shown in Table \ref{tab:exp.3datasets}, HIPSS demonstrates clear improvements over the SOTA methods across all few-shot settings on TCGA-Lung dataset. It exceeds the performance of the second best method by 1.2\%, 3.7\%, 7.8\%, and 1.9\% in 1, 2, 4, and 8-shot settings respectively.

\noindent\textbf{UBC-OCEAN (5 classes):} As demonstrated in Table \ref{tab:exp.3datasets}, performance enhancements also persist on UBC-OCEAN dataset, reporting improvements of 3.7\%, 1.8\%, 1.8\%, and 1.1\% for 2, 4, 8, and 16 shots respectively with relative to second-best method. Under the most constrained 1-shot setting, our method outperforms SOTA baseline by a significant margin of 13.8\%. 

\begin{figure}
\centering
\includegraphics[width=0.95\linewidth]{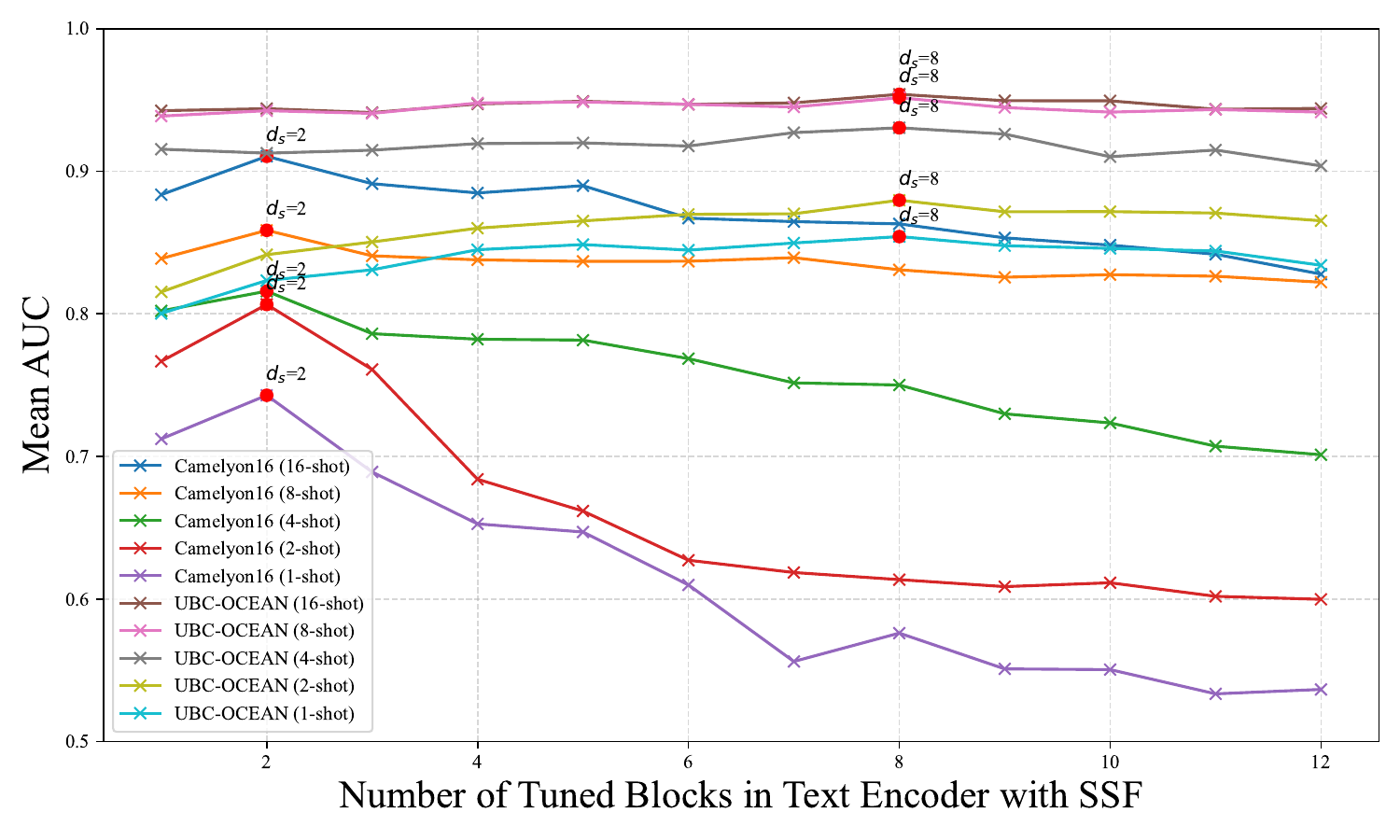}
\caption{Variation of AUC values based on the number of tuned blocks \(d_{s}\) in the text encoder with SSF. Mean AUC values are reported. The \(d_{s}\) values with the highest AUC are denoted by red dots.}
\label{fig:exp.ssf_ablation}
\end{figure}

\begin{figure}
\centering
\includegraphics[width=0.98\linewidth]{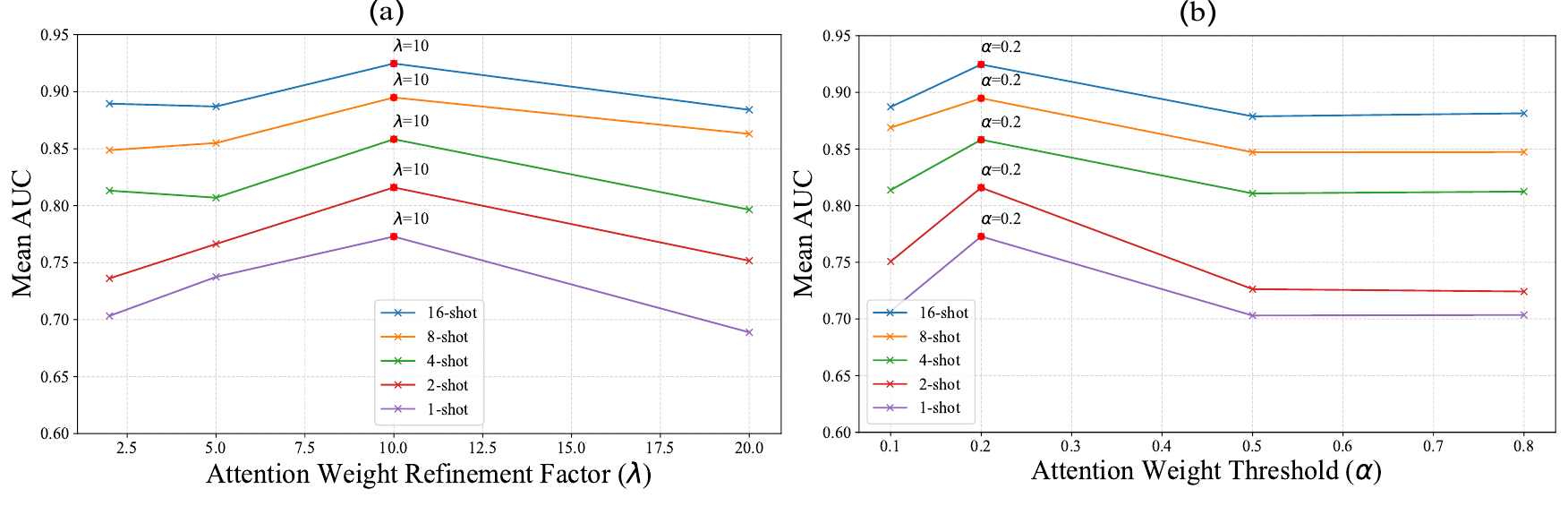}
\caption{Variation of AUC values based on \textbf{(a):} the attention weights refinement factor (\(\lambda\)) and \textbf{(b):} threshold value (\(\alpha\)). We report the mean AUC averaged across three datasets. The \(\lambda\) and \(\alpha\) values with the highest AUC are denoted by red dots.}
\label{fig:exp.Lambda_Alpha}
\end{figure}

\subsection{Ablations and Analysis}
\label{sec:ablation_and_analysis}

\begin{table}[t]
\caption{Computational efficiency comparison. Total number of trainable parameters (in millions) and average inference time per slide (in ms) are reported. The reduction compared to the second efficient FSWC method is indicated by $\downarrow$.}
\centering
\resizebox{0.95\columnwidth}{!}{
\centering
\begin{tabular}{c|cc|c}
\toprule[1.0pt]
\multirow{2}{*}{Method} & \multicolumn{2}{c|}{Training Parameters} & {Avg Inference} \\
\cmidrule{2-3}
& C16/ TCGA-Lung & UBC-OCEAN & time per slide (C16)\\
\midrule[1.0pt]
TOP \cite{top} & $0.3472$ & $0.3702$ & $58.81$ \\
FOCUS \cite{focus} & $1.3276$ & $1.3783$ & $328.49$ \\
ViLa-MIL \cite{vila-mil} & $2.6516$ & $2.6747$ & $27.30$ \\
\cellcolor{gray!20!white} HIPSS (Ours) & \cellcolor{gray!20!white} $\bm{0.2844}$ ($\downarrow 18.1\%$ ) & \cellcolor{gray!20!white} $\bm{0.3489}$ ($\downarrow 5.8\%$ ) & \cellcolor{gray!20!white} $\bm{23.85}$ ($\downarrow 12.6\%$ )\\
\bottomrule[1.0pt]
\end{tabular}
}
\label{tab:exp.efficiency}
\end{table}

\begin{table}[t]
\caption{Weakly-supervised tumor localization results on C16 Dataset as an evaluation protocol on the quality of learned feature representations. Dice coefficient is reported. Results of conventional MIL methods are cited from \cite{camil}.}
\scriptsize
\centering
\begin{tabular}{c|cc}
\toprule[1.0pt]
Method Type & Method & Dice Coefficient   \\
\midrule[1.0pt]
\multirow{5}{*}{Conventional MIL methods}
& CLAM-SB & $0.459$ \\
& TransMIL & $0.103$ \\
& DTFD-MIL & \underline{$0.525$} \\
& DSMIL & $0.259$ \\
& CAMIL & $0.515$ \\
\midrule[1.0pt]
\multirow{3}{*}{Few-shot methods}
& ABMIL \cite{abmil} & $0.164$ \\
& TOP \cite{top} & $0.203$ \\
& \cellcolor{gray!20!white} HIPSS (Ours) & \cellcolor{gray!20!white} $\bm{0.732}$  \\
\bottomrule[1.0pt]
\end{tabular}
\label{tab:exp.segmentation}
\vspace{-1.0em}
\end{table}

\begin{figure*}[t]
\centering
\includegraphics[width=0.95\textwidth]{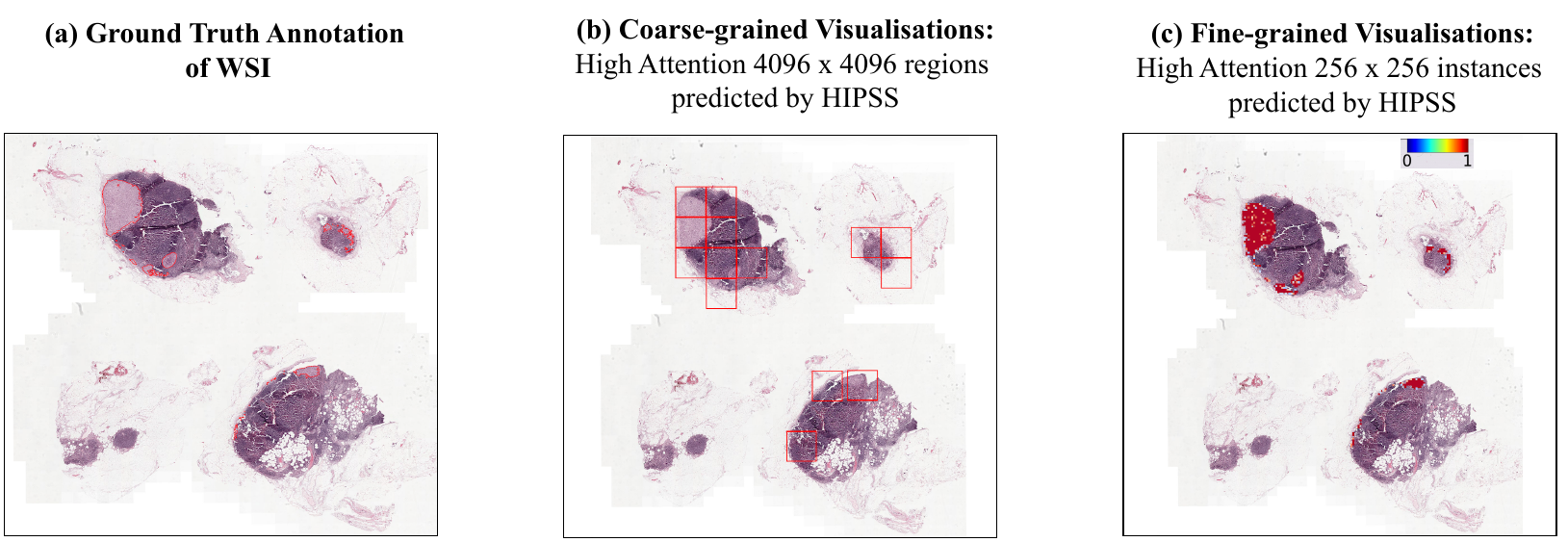}
\caption{Attention map visualizations from HIPSS compared with ground truth annotation. \textbf{(b)}: High attention regions predicted by the WSI Encoder. \textbf{(c)}: High attention instances predicted by the region encoder. Attention maps are overlaid where red indicates instances of higher attention. Quantitative results in Table \ref{tab:exp.segmentation} further demonstrates the tumor localization capabilities of our method.}
\label{fig:exp.visualization}
\end{figure*}

\textbf{Ablation Studies:} In Table \ref{tab:exp.ablations}, we show the impact of the components in HIPSS, considering hierarchical textual guidance based WSI representation learning and SSF-based prompt tuning. Strong average performance improvements of 23.5\%, 15.9\%, and 11.4\% observed under 1, 4, and 16-shot settings respectively when SSF-based prompt tuning is integrated compared to the baseline model. Additionally, an average performance gain of 3.1\% is demonstrated when the text-guided attention refinement strategy is integrated into the hierarchical attention pooling approach.

\noindent\textbf{Impact of varying tuning depth of text encoder with SSF:} Fig. \ref{fig:exp.ssf_ablation} presents the mean AUC values across C16 and UBC-OCEAN datasets as the number of tuned text encoder blocks \(d_{s}\) varies, where \(0 <d_{s}\leq  L\) and \(L=12\). A key observation is that, there is no single value of \(d_{s}\) that consistently yields optimal results across all datasets. This behavior is expected, as each dataset exhibits distinct morphological characteristics and domain shifts relative to the pre-training distribution of the VLM. However, our method achieves the peak performance for all datasets when \(d_{s} < 12\), indicating that it maintains superior performance with fewer trainable parameters.

\noindent\textbf{Impact of varying attention weight refinement factor and threshold values in soft hierarchical textual guidance:} Fig. \ref{fig:exp.Lambda_Alpha} (a) presents the impact of attention weight refinement factor ($\lambda$) on model performance, while Fig. \ref{fig:exp.Lambda_Alpha} (b) demonstrates the effect of different threshold values ($\alpha$) in hierarchical textual guidance strategy, with results averaged across three datasets. We observed that performance varies consistently across all datasets rather than being highly sensitive to parameter choices.

\noindent\textbf{Comparison with LoRA-based prompt tuning:} Table \ref{tab:exp.lora} shows the performance gains over LoRA-based prompt tuning approach. Our method achieves a mean improvement of 16.4\% across all shot settings on C16, while reducing the number of trainable parameters by 59.7\%.

\noindent\textbf{Computational efficiency analysis:} Table \ref{tab:exp.efficiency} compares the total number of trainable parameters and inference time across our method and 3 baselines. TOP \cite{top} employs CoOp-based prompt learning with contrastive alignment, while FOCUS \cite{focus} and ViLa-MIL \cite{vila-mil} combine CoOp with cross-attention, resulting in a substantially larger parameter count. Despite TOP having fewer trainable parameters, our method achieves a further reduction while maintaining SOTA performance. In addition, HIPSS reduces the average inference time per slide by 12.6\% for C16 dataset demonstrating minimal inference overhead.

\noindent\textbf{Attention map visualizations:} As qualitative analysis, we visualize the attention maps obtained by our method in Fig. \ref{fig:exp.visualization}. Since HIPSS has a hierarchical-structure aware WSI representation learning approach, we can go from coarse-grained visualizations to fine-grained visualizations.

\noindent\textbf{Tumor localization evaluation:} Though our method is trained for FSWC, it demonstrates strong weakly-supervised localization capabilities as shown in Table \ref{tab:exp.segmentation}. We report segmentation performance using dice coefficient on C16 dataset as an evaluation protocol on the quality of the learned feature representations. We utilize a simple thresholding mechanism using the attention scores for the evaluation following \cite{camil}. Our method achieves a dice coefficient of 0.732 without any localization-specific finetuning.

\section{Conclusion}
\label{sec:conclusion}

This paper proposes a new method for FSWC, which delivers a hierarchical textual guidance strategy and a SSF-based prompt tuning approach. HIPSS consistently outperforms the existing methods across diverse pathology datasets, while highlighting its computational efficiency. Furthermore, the strong weakly supervised localization capability of HIPSS highlights its potential for other downstream tasks including tumor segmentation. A key limitation is the lack of validation of LLM-generated descriptions, which could be addressed through expert-in-the-loop evaluations.

{
    \small
    \bibliographystyle{ieeenat_fullname}
    \bibliography{main}

@article{conch,
  title={A visual-language foundation model for computational pathology},
  author={Lu, Ming Y and Chen, Bowen and Williamson, Drew FK and Chen, Richard J and Liang, Ivy and Ding, Tong and Jaume, Guillaume and Odintsov, Igor and Le, Long Phi and Gerber, Georg and others},
  journal={Nature medicine},
  volume={30},
  number={3},
  pages={863--874},
  year={2024},
  publisher={Nature Publishing Group US New York}
}

@inproceedings{abmil,
  title={Attention-based deep multiple instance learning},
  author={Ilse, Maximilian and Tomczak, Jakub and Welling, Max},
  booktitle={International conference on machine learning},
  pages={2127--2136},
  year={2018},
  organization={PMLR}
}

@article{camil,
  title={CAMIL: Context-aware multiple instance learning for cancer detection and subtyping in whole slide images},
  author={Fourkioti, Olga and De Vries, Matt and Jin, Chen and Alexander, Daniel C and Bakal, Chris},
  journal={arXiv preprint arXiv:2305.05314},
  year={2023}
}

@inproceedings{dsmil,
  title={Dual-stream multiple instance learning network for whole slide image classification with self-supervised contrastive learning},
  author={Li, Bin and Li, Yin and Eliceiri, Kevin W},
  booktitle={Proceedings of the IEEE/CVF conference on computer vision and pattern recognition},
  pages={14318--14328},
  year={2021}
}

@article{fast,
  title={Fast: A dual-tier few-shot learning paradigm for whole slide image classification},
  author={Fu, Kexue and Qu, Linhao and Wang, Shuo and Xiong, Ying and Maglogiannis, Ilias and Gao, Longxiang and Wang, Manning and others},
  journal={Advances in Neural Information Processing Systems},
  volume={37},
  pages={105090--105113},
  year={2024}
}

@inproceedings{focus,
  title={Focus: Knowledge-enhanced adaptive visual compression for few-shot whole slide image classification},
  author={Guo, Zhengrui and Xiong, Conghao and Ma, Jiabo and Sun, Qichen and Feng, Lishuang and Wang, Jinzhuo and Chen, Hao},
  booktitle={Proceedings of the Computer Vision and Pattern Recognition Conference},
  pages={15590--15600},
  year={2025}
}

@article{coop,
  title={Learning to prompt for vision-language models},
  author={Zhou, Kaiyang and Yang, Jingkang and Loy, Chen Change and Liu, Ziwei},
  journal={International Journal of Computer Vision},
  volume={130},
  number={9},
  pages={2337--2348},
  year={2022},
  publisher={Springer}
}

@inproceedings{mi-zero,
  title={Visual language pretrained multiple instance zero-shot transfer for histopathology images},
  author={Lu, Ming Y and Chen, Bowen and Zhang, Andrew and Williamson, Drew FK and Chen, Richard J and Ding, Tong and Le, Long Phi and Chuang, Yung-Sung and Mahmood, Faisal},
  booktitle={Proceedings of the IEEE/CVF conference on computer vision and pattern recognition},
  pages={19764--19775},
  year={2023}
}

@article{mscpt,
  title={Mscpt: Few-shot whole slide image classification with multi-scale and context-focused prompt tuning},
  author={Han, Minghao and Qu, Linhao and Yang, Dingkang and Zhang, Xukun and Wang, Xiaoying and Zhang, Lihua},
  journal={IEEE Transactions on Medical Imaging},
  year={2025},
  publisher={IEEE}
}

@inproceedings{hipt,
  title={Scaling vision transformers to gigapixel images via hierarchical self-supervised learning},
  author={Chen, Richard J and Chen, Chengkuan and Li, Yicong and Chen, Tiffany Y and Trister, Andrew D and Krishnan, Rahul G and Mahmood, Faisal},
  booktitle={Proceedings of the IEEE/CVF conference on computer vision and pattern recognition},
  pages={16144--16155},
  year={2022}
}

@inproceedings{pemp,
  title={Pathology-knowledge enhanced multi-instance prompt learning for few-shot whole slide image classification},
  author={Qu, Linhao and Yang, Dingkang and Huang, Dan and Guo, Qinhao and Luo, Rongkui and Zhang, Shaoting and Wang, Xiaosong},
  booktitle={European conference on computer vision},
  pages={196--212},
  year={2024},
  organization={Springer}
}

@article{trans-mil,
  title={Transmil: Transformer based correlated multiple instance learning for whole slide image classification},
  author={Shao, Zhuchen and Bian, Hao and Chen, Yang and Wang, Yifeng and Zhang, Jian and Ji, Xiangyang and others},
  journal={Advances in neural information processing systems},
  volume={34},
  pages={2136--2147},
  year={2021}
}

@inproceedings{vila-mil,
  title={Vila-mil: Dual-scale vision-language multiple instance learning for whole slide image classification},
  author={Shi, Jiangbo and Li, Chen and Gong, Tieliang and Zheng, Yefeng and Fu, Huazhu},
  booktitle={Proceedings of the IEEE/CVF Conference on Computer Vision and Pattern Recognition},
  pages={11248--11258},
  year={2024}
}

@inproceedings{dtfd-mil,
  title={Dtfd-mil: Double-tier feature distillation multiple instance learning for histopathology whole slide image classification},
  author={Zhang, Hongrun and Meng, Yanda and Zhao, Yitian and Qiao, Yihong and Yang, Xiaoyun and Coupland, Sarah E and Zheng, Yalin},
  booktitle={Proceedings of the IEEE/CVF conference on computer vision and pattern recognition},
  pages={18802--18812},
  year={2022}
}

@article{top,
  title={The rise of ai language pathologists: Exploring two-level prompt learning for few-shot weakly-supervised whole slide image classification},
  author={Qu, Linhao and Fu, Kexue and Wang, Manning and Song, Zhijian and others},
  journal={Advances in Neural Information Processing Systems},
  volume={36},
  pages={67551--67564},
  year={2023}
}

@inproceedings{clip,
  title={Learning transferable visual models from natural language supervision},
  author={Radford, Alec and Kim, Jong Wook and Hallacy, Chris and Ramesh, Aditya and Goh, Gabriel and Agarwal, Sandhini and Sastry, Girish and Askell, Amanda and Mishkin, Pamela and Clark, Jack and others},
  booktitle={International conference on machine learning},
  pages={8748--8763},
  year={2021},
  organization={PmLR}
}

@article{plip,
  title={A visual--language foundation model for pathology image analysis using medical twitter},
  author={Huang, Zhi and Bianchi, Federico and Yuksekgonul, Mert and Montine, Thomas J and Zou, James},
  journal={Nature medicine},
  volume={29},
  number={9},
  pages={2307--2316},
  year={2023},
  publisher={Nature Publishing Group US New York}
}

@article{quilt,
  title={Quilt-1m: One million image-text pairs for histopathology},
  author={Ikezogwo, Wisdom and Seyfioglu, Saygin and Ghezloo, Fatemeh and Geva, Dylan and Sheikh Mohammed, Fatwir and Anand, Pavan Kumar and Krishna, Ranjay and Shapiro, Linda},
  journal={Advances in neural information processing systems},
  volume={36},
  pages={37995--38017},
  year={2023}
}

@article{camelyon16,
  title={Diagnostic assessment of deep learning algorithms for detection of lymph node metastases in women with breast cancer},
  author={Bejnordi, Babak Ehteshami and Veta, Mitko and Van Diest, Paul Johannes and Van Ginneken, Bram and Karssemeijer, Nico and Litjens, Geert and Van Der Laak, Jeroen AWM and Hermsen, Meyke and Manson, Quirine F and Balkenhol, Maschenka and others},
  journal={Jama},
  volume={318},
  number={22},
  pages={2199--2210},
  year={2017},
  publisher={American Medical Association}
}

@misc{ubc,
    author = {Ali Bashashati and Hossein Farahani and OTTA Consortium and Anthony Karnezis and Ardalan Akbari and Sirim Kim and Ashley Chow and Sohier Dane and Allen Zhang and Maryam Asadi},
    title = {UBC Ovarian Cancer Subtype Classification and Outlier Detection (UBC-OCEAN)},
    year = {2023},
    howpublished = {\url{https://kaggle.com/competitions/UBC-OCEAN}},
    note = {Kaggle}
}

@article{clam,
  title={Data-efficient and weakly supervised computational pathology on whole-slide images},
  author={Lu, Ming Y and Williamson, Drew FK and Chen, Tiffany Y and Chen, Richard J and Barbieri, Matteo and Mahmood, Faisal},
  journal={Nature biomedical engineering},
  volume={5},
  number={6},
  pages={555--570},
  year={2021},
  publisher={Nature Publishing Group UK London}
}

@article{ssf,
  title={Scaling \& shifting your features: A new baseline for efficient model tuning},
  author={Lian, Dongze and Zhou, Daquan and Feng, Jiashi and Wang, Xinchao},
  journal={Advances in Neural Information Processing Systems},
  volume={35},
  pages={109--123},
  year={2022}
}

@article{mil-survery1,
  title={Deep neural network models for computational histopathology: A survey},
  author={Srinidhi, Chetan L and Ciga, Ozan and Martel, Anne L},
  journal={Medical image analysis},
  volume={67},
  pages={101813},
  year={2021},
  publisher={Elsevier}
}

@article{mil-survey2,
  title={Artificial intelligence and computational pathology},
  author={Cui, Miao and Zhang, David Y},
  journal={Laboratory Investigation},
  volume={101},
  number={4},
  pages={412--422},
  year={2021},
  publisher={Elsevier}
}

@article{mil-survey3,
  title={Label-efficient deep learning in medical image analysis: Challenges and future directions},
  author={Jin, Cheng and Guo, Zhengrui and Lin, Yi and Luo, Luyang and Chen, Hao},
  journal={arXiv preprint arXiv:2303.12484},
  year={2023}
}

@article{cpath-survey1,
  title={Computational pathology: a survey review and the way forward},
  author={Hosseini, Mahdi S and Bejnordi, Babak Ehteshami and Trinh, Vincent Quoc-Huy and Chan, Lyndon and Hasan, Danial and Li, Xingwen and Yang, Stephen and Kim, Taehyo and Zhang, Haochen and Wu, Theodore and others},
  journal={Journal of Pathology Informatics},
  volume={15},
  pages={100357},
  year={2024},
  publisher={Elsevier}
}

@article{fewshot1,
  title={Towards better understanding and better generalization of few-shot classification in histology images with contrastive learning},
  author={Yang, Jiawei and Chen, Hanbo and Yan, Jiangpeng and Chen, Xiaoyu and Yao, Jianhua},
  journal={arXiv preprint arXiv:2202.09059},
  year={2022}
}

@article{vlm-survey,
  title={Vision-language models for vision tasks: A survey},
  author={Zhang, Jingyi and Huang, Jiaxing and Jin, Sheng and Lu, Shijian},
  journal={IEEE transactions on pattern analysis and machine intelligence},
  volume={46},
  number={8},
  pages={5625--5644},
  year={2024},
  publisher={IEEE}
}

@inproceedings{align,
  title={Scaling up visual and vision-language representation learning with noisy text supervision},
  author={Jia, Chao and Yang, Yinfei and Xia, Ye and Chen, Yi-Ting and Parekh, Zarana and Pham, Hieu and Le, Quoc and Sung, Yun-Hsuan and Li, Zhen and Duerig, Tom},
  booktitle={International conference on machine learning},
  pages={4904--4916},
  year={2021},
  organization={PMLR}
}

@article{filip,
  title={Filip: Fine-grained interactive language-image pre-training},
  author={Yao, Lewei and Huang, Runhui and Hou, Lu and Lu, Guansong and Niu, Minzhe and Xu, Hang and Liang, Xiaodan and Li, Zhenguo and Jiang, Xin and Xu, Chunjing},
  journal={arXiv preprint arXiv:2111.07783},
  year={2021}
}

@article{annotationeffmil,
  title={Annotation-efficient deep learning for breast cancer whole-slide image classification using tumour infiltrating lymphocytes and slide-level labels},
  author={Perera, Rashindrie and Savas, Peter and Senanayake, Damith and Salgado, Roberto and Joensuu, Heikki and O’Toole, Sandra and Li, Jason and Loi, Sherene and Halgamuge, Saman},
  journal={Communications Engineering},
  volume={3},
  number={1},
  pages={104},
  year={2024},
  publisher={Nature Publishing Group UK London}
}

@inproceedings{fewshotdiscriminative,
  title={Discriminative sample-guided and parameter-efficient feature space adaptation for cross-domain few-shot learning},
  author={Perera, Rashindrie and Halgamuge, Saman},
  booktitle={Proceedings of the IEEE/CVF Conference on Computer Vision and Pattern Recognition},
  pages={23794--23804},
  year={2024}
}

@article{adam,
  title={Adam: A method for stochastic optimization},
  author={Kingma, Diederik P and Ba, Jimmy},
  journal={arXiv preprint arXiv:1412.6980},
  year={2014}
}

@article{MAPLE,
  title={MAPLE: Multi-scale Attribute-enhanced Prompt Learning for Few-shot Whole Slide Image Classification},
  author={Zhou, Junjie and Shao, Wei and Yue, Yagao and Mu, Wei and Wan, Peng and Zhu, Qi and Zhang, Daoqiang},
  journal={arXiv preprint arXiv:2509.25863},
  year={2025}
}

@article{benchmarkFM,
  title={Benchmarking foundation models as feature extractors for weakly supervised computational pathology},
  author={Neidlinger, Peter and El Nahhas, Omar SM and Muti, Hannah Sophie and Lenz, Tim and Hoffmeister, Michael and Brenner, Hermann and van Treeck, Marko and Langer, Rupert and Dislich, Bastian and Behrens, Hans Michael and others},
  journal={Nature biomedical engineering},
  pages={1--11},
  year={2025},
  publisher={Nature Publishing Group UK London}
}
}


\end{document}